\pdfoutput=1

\documentclass[11pt]{article}

\usepackage{EACL2023}

\usepackage{times}
\usepackage{latexsym}

\usepackage[T1]{fontenc}

\usepackage[utf8]{inputenc}

\usepackage{microtype}

\usepackage{inconsolata}

\usepackage{lineno,hyperref}
\usepackage{xurl}
\usepackage{graphicx}
\usepackage{todonotes,comment,pbox,multirow,arabtex,utf8,pifont,graphics,amsmath,textcomp,subcaption,array,float,adjustbox,subcaption,caption,pbox,amssymb,amsmath,booktabs}
\usepackage{algorithm}
\usepackage{algorithmic}
\usepackage{tabularx}
\usepackage{colortbl}
\usepackage{xifthen}

\newcommand\randmap{{\sc MapRand}}
\newcommand\predmap{{\sc MapPred}}
\newcommand\randmapint{{\sc MapRand$_{1-1}$}}
\newcommand\randmapgdf{{\sc MapRand$_{n-n}$}}
\newcommand\predmapint{{\sc MapPred$_{1-1}$}}
\newcommand\predmapgdf{{\sc MapPred$_{n-n}$}}

\newcommand\dict{{\sc Dictionary}}

\title{Investigating Lexical Replacements \\ for Arabic-English Code-Switched Data Augmentation}

\author{Injy Hamed,$^{1,2}$ Nizar Habash,$^{1}$ Slim Abdennadher,$^{3}$ Ngoc Thang Vu$^{2}$ \\
  $^1$Computational Approaches to Modeling Language Lab, 
  New York University Abu Dhabi \\
  $^2$Institute for Natural Language Processing, University of Stuttgart\\
  $^3$Informatics and Computer Science, The German International University in Cairo\\
  \texttt{injy.hamed@nyu.edu}
  }

\begin{document}
\maketitle
\begin{abstract}
Data sparsity is a main problem hindering the development of code-switching (CS) NLP systems. In this paper, we investigate data augmentation techniques for synthesizing dialectal Arabic-English CS text. We perform lexical replacements using word-aligned parallel corpora where CS points are either randomly chosen or learnt using a sequence-to-sequence model. We compare these approaches against dictionary-based replacements. We assess the quality of the generated sentences through human evaluation and evaluate the effectiveness of data augmentation on machine translation (MT), automatic speech recognition (ASR), and speech translation (ST) tasks. Results show that using a predictive model results in more natural CS sentences compared to the random approach, as reported in human judgements. In the downstream tasks, despite the random approach generating more data, both approaches perform equally (outperforming dictionary-based replacements). Overall, data augmentation achieves 34\% improvement in perplexity, 5.2\% relative improvement on WER for ASR task, +4.0-5.1 BLEU points on MT task, and +2.1-2.2 BLEU points on ST over a baseline trained on available data without augmentation.
 \end{list}
\end{abstract}

\setcode{utf8}

\section{Introduction}
Code-switching (CS) is the alternation of language in text or speech. CS can occur at the levels of sentences (inter-sentential CS), words (intra-sentential CS/code-mixing), and morphemes (intra-word CS/morphological CS). 
Given that CS data is scarce and that collecting such data is expensive and time-consuming, data augmentation serves as a successful solution 
for alleviating data sparsity. 

In this paper, we investigate lexical replacements for augmenting CS dialectal Arabic-English data. Researchers have investigated approaches that do not require parallel data, including translating source words into target language with the use of dictionaries \cite{TKJ21}, machine translation \cite{LV20}, and word embeddings \cite{SOW+21}, as well as relying on parallel data and performing substitutions of words/phrases using alignments \cite{MLJ+19,AGE+21,GVS21}. 
As will be discussed in Section \ref{sec:related_work}, most of the previous studies on this front have focused on one augmentation technique without exploring others, or reported results using only one type of word alignments configuration, or evaluated effectiveness of augmentation on only one downstream task. 

We attempt to provide a comprehensive study where we systematically explore the use of neural-based models to decide on CS points for performing replacements using word-aligned parallel corpora versus randomly-chosen CS points, along with the interaction of different alignment configurations. We compare these approaches against dictionary-based replacements. We provide a rigorous evaluation of the different settings, where we assess the quality of the generated CS sentences through human evaluation as well as
the impact on language modeling (LM), automatic speech recognition (ASR), machine translation (MT), and speech translation (ST) tasks. 

Our human evaluation study shows that for the purpose of generating high-quality CS sentences, learning to predict CS points and integrating this information in the augmentation process improves the quality of generated sentences. 
On the downstream tasks, we report that performing alignment-based replacement outperforms dictionary-based replacement. 
For alignment-based replacement, utilizing a predictive model to decide on where CS points should occur as opposed to replacing at random positions both lead to similar results for ASR, MT, and ST tasks. For both approaches, we investigate different word alignment configurations, and we report that performing segment replacements using symmetrized alignments outperforms word-replacements using intersection alignments on both human evaluation and extrinsic evaluation. We also investigate controlling the amount of generated data, to eliminate the effect of random producing more data over the predictive model. Under the constrained condition, using a predictive model outperforms the random approach on the MT task.


In this work, we tackle the following research questions (RQs):
\begin{itemize}
\item \textbf{RQ1:} Can a model learn to predict CS points using limited amount of CS data?
\item \textbf{RQ2:} Can this information be used to generate more natural synthetic CS data?
\item \textbf{RQ3:} Would higher quality of synthesized CS data necessarily reflect in performance improvements in downstream tasks?
\end{itemize}

\section{Related Work}
\label{sec:related_work}
Most of the work done for CS data augmentation has been focused on LM, mostly for ASR. Several techniques have been proposed based on linguistic theories \cite{PBC+18,LYL19,hussein2023textual}, heuristics \cite{SWY+11,VLW+12,KAT+21}, neural networks \cite{CCL18,WMW+18,WMW+19,LV20}, and MT \cite{TKJ21}. CS data augmentation has been less investigated for MT. Previous work has mainly involved lexical replacements \cite{MLJ+19,SZY+19,AGE+21,GVS21,XY21} and back translation \cite{KFC+21}. In this section, we discuss previous work that we find closest to ours. 


\newcite{hussein2023textual} generated synthetic CS Arabic-English text based on the equivalence constraint (EC) theory \cite{Pop00} using the GCM tool \cite{RSG+21}, as well as random lexical replacements. It was shown that while relying on the EC theory generates more natural CS sentences, as shown in human evaluation, using lexical replacements outperforms the linguistic-based approach on LM and ASR tasks. 

In the direction of lexical replacements, 
\newcite{AGE+21} generated synthetic CS Hindi-English sentences by replacing all source words (except for stopwords) by the corresponding target words using 1-1 alignments, achieving improvements on MT task. 
\newcite{GVS21} trained a neural-based model to predict CS points on monolingual source text. Using 1-n alignments, the source word is replaced by the aligned word(s). They evaluate their approach against unigram and bigram random replacements, and test its effectiveness on MT task for CS Hindi-English. \newcite{XY21} use data augmentation for MT task for CS Spanish-English and French-English. Symmetrized alignments are used to identify small aligned phrases (minimal alignment units) and phrase replacements are performed randomly. We also notice that in literature, human evaluation of generated CS data is mainly used to evaluate the synthetic data produced by the best model, rather than comparing different techniques. Such a comparison was provided by \newcite{PC21}, where a large-scale human evaluation was presented comparing different linguistic-driven and lexical replacement techniques. However, the study was focused on human evaluation without exploring the effectiveness of those techniques on downstream tasks.

\section{Data Augmentation}
\begin{figure}[t]
    \centering
    \includegraphics[width=\columnwidth]{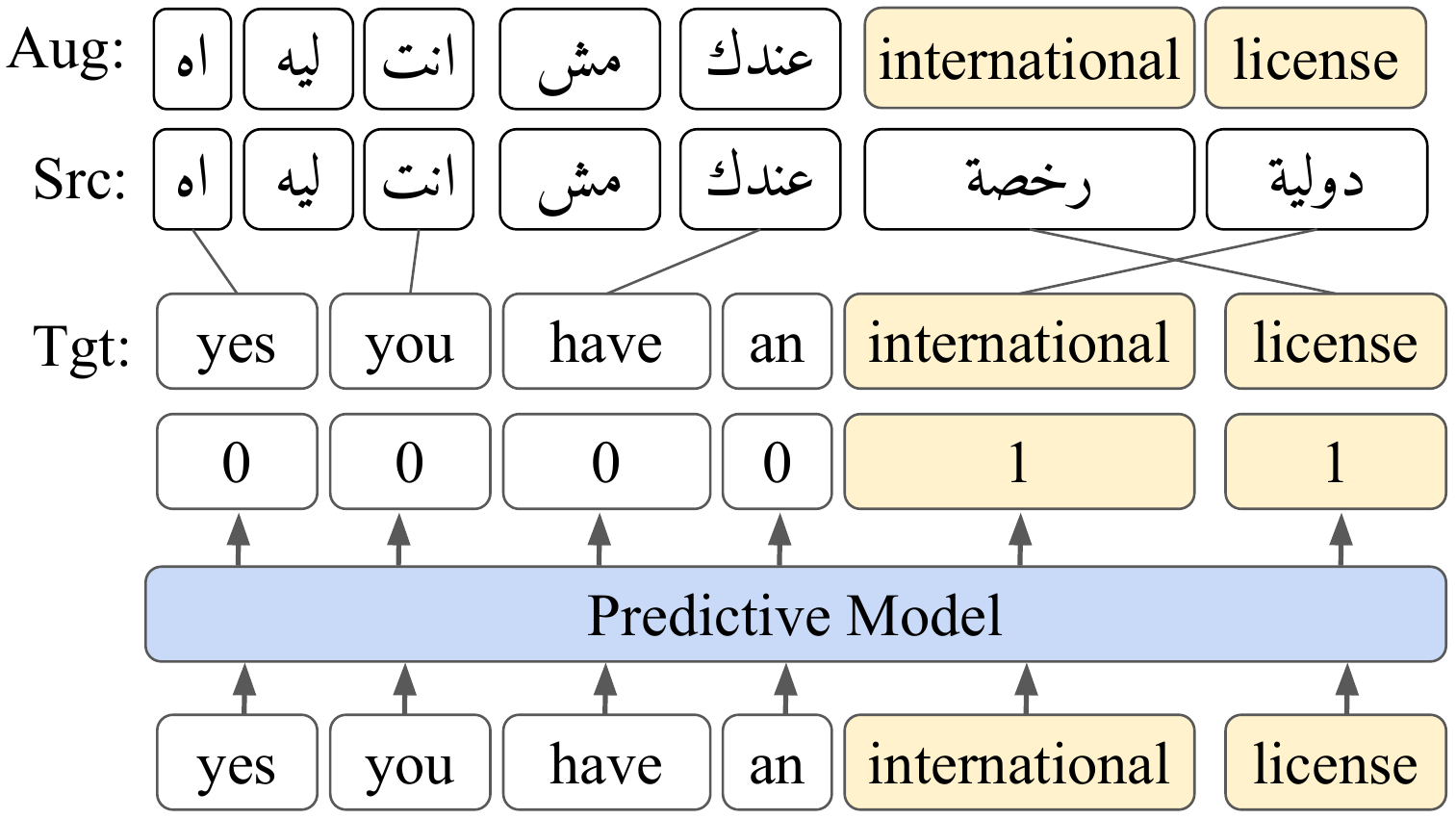}
    \caption{Data augmentation process.}
    \label{fig:augmentation}
\end{figure}

For generating synthetic CS data, we investigate the use of word-aligned parallel sentences as well as dictionary-based replacements. In the latter approach, monolingual Arabic sentences are augmented by replacing words at random locations with their English glossary entry. 
In the former approach, utilizing monolingual Arabic-English parallel corpora, we inject words from the target side to the source side, where replacements are performed at random locations or using a CS point predictive model. As shown in Figure \ref{fig:augmentation}, the augmentation process consists of two main steps: (1) CS point prediction: identifying the target words to be borrowed, and (2) CS generation: performing the replacements. In Sections \ref{sec:CS_prediction} and \ref{sec:CS_generation}, we will elaborate on the methodology for both steps.

\label{sec:data_augmentation}

\begin{table*}[t]
\centering
\setlength{\tabcolsep}{2pt}
\begin{tabular}{ |l| l|}
\cline{2-2}
\multicolumn{1}{c}{}&\multicolumn{1}{|c|}{\textbf{Examples}}\\\hline
Src&
\multicolumn{1}{r|}{
.
\<شوية>
\textcolor{blue}{academic life}
\<ال>
\textcolor{blue}{i love}
\<في فترة فجربت الموضوع ف>
\textcolor{blue}{i was a junior ta}
\<و>
$\leftarrow$}\\
Tgt& and \textcolor{blue}{\underline{i was a junior ta}} for a period of time so i have tried this and \textcolor{blue}{\underline{i love}} the \textcolor{blue}{\underline{academic life}} a bit .\\
Output& {} {} 0 {} {}\textcolor{blue}{1} {} { } \textcolor{blue}{1}  {} \textcolor{blue}{1} { } { }\textcolor{blue}{1}  { } { } {} \textcolor{blue}{1} { } {}0 { }0  { } { } 0  { } { } 0 { } { }  0  { } {} 0{} {}0  { } {} 0  { } { } { }0  { } { } 0  { } { } 0  { } \textcolor{blue}{1} { } \textcolor{blue}{1}  { } { } 0  { } { } { } { } { }\textcolor{blue}{1}  { } { } { } { }\textcolor{blue}{1} { } {}0 {} {}0 {}{}0\\\hline
Src&
\multicolumn{1}{r|}{
\<زي دي اساسا>
\textcolor{blue}{city}
\<اني اشوف>
\textcolor{blue}{expect}
\<ماكنتش م>
$\leftarrow$}\\
Tgt& i wasn 't \textcolor{blue}{\underline{expecting}} to see such a \textcolor{blue}{\underline{city}} in the first place .\\
Output& 0 {} { } 0 {} {} {} 0  { } { } { } { }\textcolor{blue}{1} { } { } { } {} 0 { } 0 { } { } 0 { } 0 { }{ }\textcolor{blue}{1} { } { }0 
  { } 0  { } { }0  { } { } { } 0  { } 0\\\hline
\end{tabular}
\caption{Example showing the matching algorithm output for source and target sentences. The matched words on the target side are underlined. The arrows show the sentence starting direction, as Arabic is read right to left.}
\label{table:matchingAlg_examples}
\end{table*}

\subsection{CS Point Prediction}
\label{sec:CS_prediction}
Similar to \newcite{GVS21}, we model the task of CS point prediction as a sequence-to-sequence classification task. The neural network takes as input the word sequence $x=\lbrace x_1, x_2, .. , x_N \rbrace$, where ${N}$ is the length of the input sentence. The network outputs a sequence $y=\lbrace y_1, y_2, .. , y_N \rbrace$, where $y_n\in \lbrace$1,0$\rbrace$ represents whether the word $x_n$ is to be code-switched or not. 
We learn CS points using ArzEn-ST corpus \cite{hamed2022arzenST}, which contains CS Egyptian Arabic-English sentences and their English translations. We then utilize the learnt CS model to augment a large number of monolingual Arabic-English parallel sentences by inserting the tagged words on the (English) target side into the (Egyptian Arabic) source side.

In order to learn CS points, the neural network needs to take as input monolingual sentences from either the source or target sides, along with tags representing whether this word should be code-switched or not. In \newcite{GVS21}, the authors generated synthetic monolingual sentences from CS sentences by translating CS segments to the source language, and then learning CS points on the source side. While this approach seems more intuitive, CS segments abide by the grammatical rules of the embedded language, thus direct translation of embedded words would result in sentences having incorrect structures in the matrix language in case of syntactic divergence, which is present between Arabic and English. 
Instead, we opt to learn CS points on the target side. 
This approach provides another advantage, as English is commonly used in CS, having the predictive model trained on English as opposed to the primary language (which could be low-resourced) allows for the use of available resources such as pretrained LMs.

The challenge in this approach is identifying the words on the target side which correspond to the CS words on the source side. Relying on the translators to perform this annotation task is costly, time consuming, and error-prone.\footnote{We have tried this annotation task for ArzEn-ST and only 72\% of the CS words got annotated.} Relying on word alignments is also not optimal, where only 83\% of CS words in ArzEn-ST train set were matched using intersection alignment. Recall could increase using a less strict alignment approach, but would be at the risk of less accurate matches. Therefore, we develop a matching algorithm that 
is based on the following idea: if a CS segment occurs ${x}$ times in the source and target sentences, then we identify these segments as matching segments. We match segments starting with the longest segments (and sub-segments) first. When matching words, we check their categorial variation \cite{HD03} as well as stems to match words having slight modifications in translation.\footnote{
In case $|matches_{tgt}|>|matches_{src}|$, we first rely on alignments to make the decision, achieving 99.6\% matches on ArzEn-ST train set, then we randomly pick matched target segments to cover the number of matches on the source side in order to 
increase recall. 
} 
This matching algorithm provides a language-agnostic approach to identify words on the target side that are code-switched segments on the source side.\footnote{Code available: \url{http://arzen.camel-lab.com/}} Examples of algorithm output are shown in Table \ref{table:matchingAlg_examples}, where it is seen that \textit{expect} and \textit{expecting} are matched as a result of the categorial variation check.



\subsection{CS Generation}
\label{sec:CS_generation}
\begin{figure}[t]
    \centering
    \includegraphics[width=0.95\columnwidth]{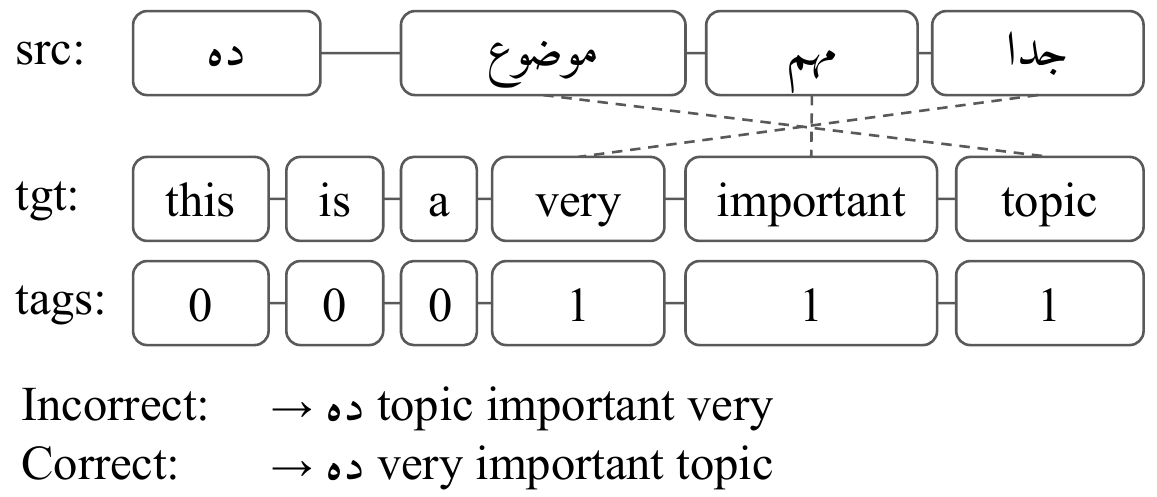}
    \caption{Data augmentation under the Continuity Constraint.}
    \label{fig:continuity_constraint}
\end{figure}

After identifying the target words to be embedded into the source side, we rely on alignments using GIZA++ \cite{CV07} to perform the replacements. 
While direct replacements can be performed in the case of single word switches, in the case of replacing multiple consecutive words, direct word replacements would produce incorrect CS structures in the case of syntactic divergence. 
In the case of Arabic-English, this is particularly evident for adjectival phrases. 
Accordingly, when performing word replacements, we maintain the same order of 
consecutive English words, 
which we refer to as the ``Continuity Constraint''. In Figure~\ref{fig:continuity_constraint}, the importance of applying this constraint is illustrated. Without such a constraint, the generated sentence outlined in Figure \ref{fig:continuity_constraint} would follow the Arabic syntactic structure resulting in ``\<ده> topic important very'' (\textit{this [is a] topic important very}).

\begin{figure}[t]
    \centering
    \includegraphics[width=\columnwidth]{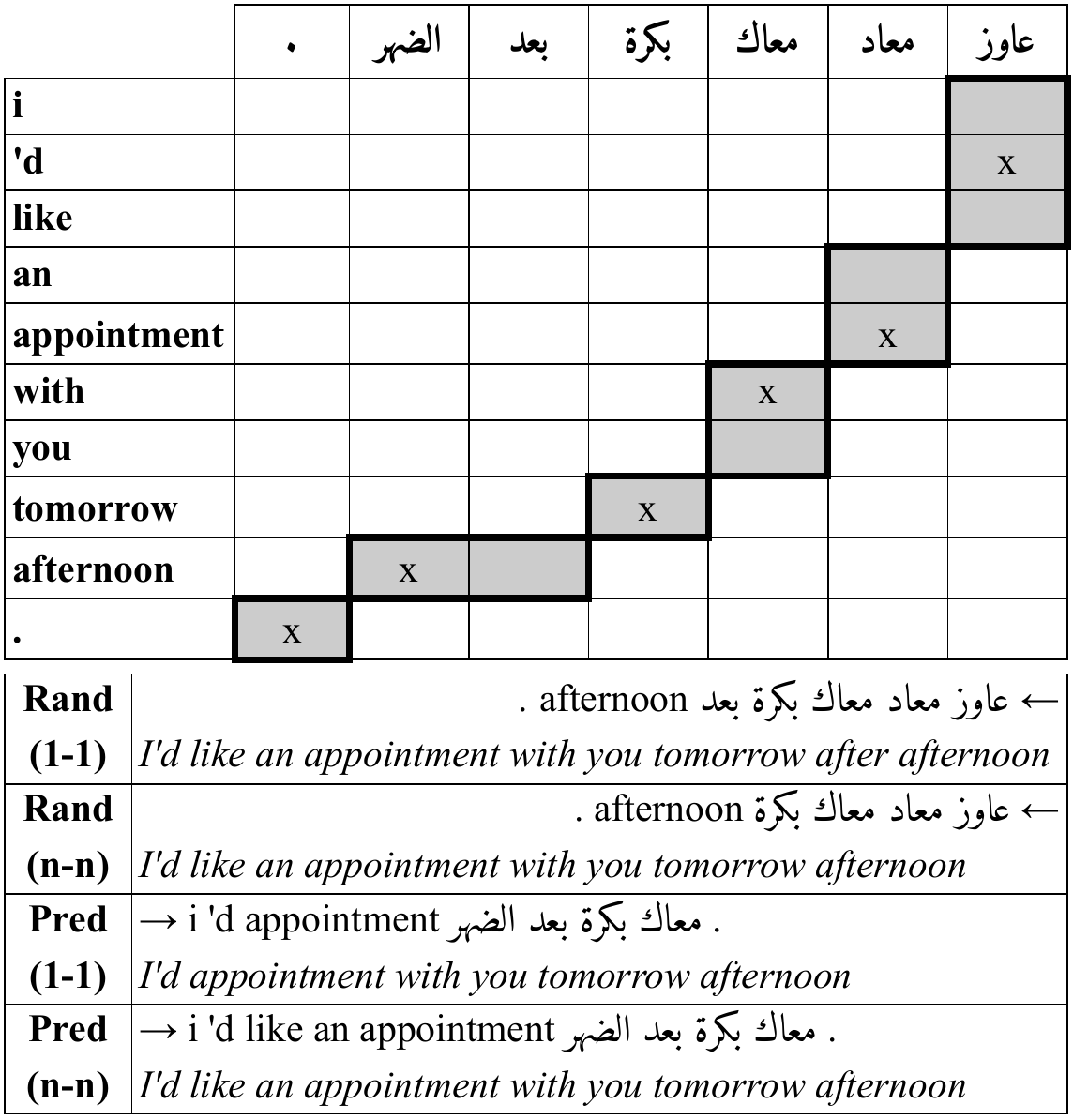}
    \caption{Example showing 1-1 and n-n alignments. Intersection alignments are marked with `x' and the grow-diag-final alignments are highlighted. We show the generated sentences with translations for each setup.
    }
    \label{fig:alignments_replacement}
\end{figure}

When performing replacements, we investigate the use of intersection alignments as well as grow-diag-final alignments.\footnote{We experiment with relying on alignments trained on word space only, stem space only, and the merge of both alignments, where for intersection alignments, we first rely on the alignments obtained in stem space, and add remaining alignments obtained from word space, such that 1-1 alignments are retained, and for grow-diag-final alignments, we take the union of alignments in both spaces. We find that merging alignments in both spaces achieves higher alignment coverage as well as better results in extrinsic tasks. Therefore, we will only be presenting the results using the merged alignments.} While intersection alignment provides high precision, relying on 1-1 alignments is not always correct, as an Arabic word can map to multiple English words and vice versa. Therefore, we investigate the use of grow-diag-final (symmetrized) alignments to identify aligned segments. The aligned segments consist of pairs of the minimal number of consecutive words (S,T) where all words in source segment (S) are aligned to one or more words in target segment (T) and are not aligned to any other words outside (T), with the same constraints applying in the opposite (target-source) direction. 
Afterwards, for each English word receiving a positive CS tag, the whole target segment containing this word replaces the aligned source segment. Throughout the paper, we will refer to the two approaches as using 1-1 and n-n alignments.
In Figure \ref{fig:alignments_replacement}, we present an example showing the results of augmentation using predictive CS models versus random CS point prediction along with using  1-1 or n-n alignments. 

\subsection{Augmentation Approaches}
We investigate the following approaches:
\paragraph{\dict:} We randomly pick ${x}$ source words and replace them with an English glossary entry using MADAMIRA \cite{PAD+14}. We set ${x}$ to 19\% of the source words, where this number is chosen based on the percentage of English words in CS sentences in ArzEn-ST train set, given that we would like to mimic natural CS behaviour. 

\paragraph{\randmap:} We randomly pick ${x}$ target words having source-target intersection alignments. We set ${x}$ to 19\% of the source words. 
We use word and segment replacements, where the models are referred to as \randmapint~and \randmapgdf.
\paragraph{\predmap:} We fine-tune pretrained mBERT model using NERDA framework \cite{KN21} to predict the target words to be injected into the source side.\footnote{We maintain the original tokenization of the input text, where we project further tokenization performed on the output into the original tokenization.} 
We use 1-1 and n-n alignments to perform replacements, where the models are referred to as \predmapint~and \predmapgdf.\footnote{For training the predictive models, we also tried using BERT models, which gave slightly lower results.} For finetuning mBERT, we set the epochs to 5, drop-out rate to 0.1, warmup steps to 500, batch size to 13, and learning rate to 0.0001. 
\section{Experiments}
\label{sec:experimental_setup}
\subsection{Data}
\label{sec:data_preparation}
We use ArzEn-ST corpus \cite{hamed2022arzenST} as our CS corpus. The corpus contains English translations of an Egyptian Arabic-English code-switched speech corpus \cite{HVA20} that is gathered through informal interviews with bilingual speakers. 
The corpus is divided into train, dev, and test sets having $3.3k$, $1.4k$, and $1.4k$ sentences (containing $2.2k$, $0.9k$, and $0.9k$ CS sentences), respectively. We follow the same data splits. 
In Appendix~\ref{sec:appendix-arzenExamples}, we provide an overview of ArzEn-ST corpus.

We also utilize the following Egyptian Arabic-English parallel corpora: Callhome Egyptian Arabic-English Speech Translation
Corpus \cite{GKA+97,LDC2002T38,LDC2002T39,KCC+14}, LDC2012T09 \cite{ZMD+12}, 
LDC2017T07 \cite{LDC2017T07}, 
LDC2019T01 \cite{LDC2019T01}, 
LDC2021T15 \cite{LDC2021T15}, and MADAR \cite{BHS18}. The corpora contain $308k$ monolingual parallel sentences as well as $15k$ CS parallel sentences. We use the same data splits as defined for each corpus. For corpora with no defined data  splits, we use the guidelines provided in \cite{DHR+13}. Data preprocessing for ArzEn-ST and the parallel corpora is discussed in Appendix~\ref{sec:appendix_data_preprocessing}.

\paragraph{Data Augmentation:} For data augmentation, we use the monolingual parallel sentences and augment them into CS parallel sentences. For the CS point predictive model, we use the CS sentences in ArzEn-ST train and dev sets for training and development, respectively.

\paragraph{MT:} The MT baseline system is trained on ArzEn-ST train set, in addition to the $308k$ monolingual parallel sentences. In the augmentation experiments, we add the augmented sentences to the baseline training data. For development and testing, we use ArzEn-ST dev and test sets.

\paragraph{ASR:} The ASR baseline system is trained on the following Egyptian Arabic data: ArzEn speech corpus \cite{HVA20}, Callhome \cite{GKA+97}, and MGB-3 \cite{AVR17}. A subset of 5-hours was used from each of Librispeech \cite{PCP+15} (English) and MGB-2 \cite{ABG+16} (MSA), where adding more data from these corpora deteriorated the ASR performance \cite{HDL+22}. 
The LM baseline model is trained on corpora transcriptions. 
For the LM models using augmented data, we append the augmented data to those transcriptions. For development and testing, we use ArzEn-ST dev and test sets. 

As an extra experiment, we compare the performance of the systems relying on synthetic CS data versus using available real CS data. For MT, we use the $15k$ CS parallel sentences in addition to the baseline data. For ASR rescoring, we train the LM on the baseline data in addition to 117,844 code-switched sentences collected from social media platforms \cite{HZE+19}. We denote these experiments as \textit{ExtraCS} in the results.

\subsection{Machine Translation System}
We train a Transformer model using Fairseq \cite{OEB+19} on a single GeForce RTX 3090 GPU. We use the hyperparameters from the FLORES benchmark for low-resource machine translation \cite{GCO+19}.\footnote{We follow \cite{gaser2022exploring}, where it was shown that FLORES hyperparameters outperform \citet{vaswani2017attention} using the same datasets.} The hyperparameters are given in Appendix \ref{sec:appendix-MThyperparameters}. 
We use a BPE model trained jointly on source and target sides with a vocabulary size of $16k$ (which outperforms $1,3,5,8,32,64k$).\footnote{For the \textit{ExtraCS} experiment, we use a vocabulary size of $8k$, which outperforms $16k$ and $32k$.} The BPE model is trained using Fairseq with character\_coverage set to $1.0$. 

\subsection{Automatic Speech Recognition System}
We train a joint CTC/attention based E2E ASR system using ESPnet \cite{WHK+18}. The encoder and decoder consist of 12 and 6 Transformer blocks with 4 heads, feed-forward inner dimension 2048 and attention dimension 256. The CTC/attention weight $(\lambda_1)$ is set to 0.3. SpecAugment \cite{PCZ+19} is applied for data augmentation. For LM, the RNNLM consists of 1 LSTM layer with 1000 hidden units and is trained for 20 epochs. For decoding, the beam size is 20 and the CTC weight is 0.2.

\subsection{Speech Translation System}
We build a cascaded ST system using the 
ASR and MT models. We opt for a cascaded system over an end-to-end system due to the limitation of available resources to build an end-to-end system, in addition to the fact that cascaded systems have shown to outperform end-to-end systems in low-resource settings \cite{DMV21}.
\section{Results}
\label{sec:results}

In order to evaluate our augmentation techniques, we provide intrinsic evaluation, extrinsic evaluation, as well as human evaluation.\footnote{The MT models require around 4 hours for training. The ASR system required around 48 hours for training, as well as 6 hours for ASR rescoring. The CS predictive model using mBERT required around 10 hours for inference.} 
According to human evaluation, the synthetic data generated using a CS predictive model is perceived as more natural. However, our extrinsic evaluation shows that both aligned-based approaches (random replacements and relying on a predictive model) perform equally on downstream tasks. We observe that using a predictive model generates less data than the random approach. When controlling for size, we observe that using a predictive model brings improvements on the MT task. Both aligned-based approaches outperform dictionary-based replacements on human evaluation and extrinsic evaluation. Regarding the effect of word alignment configurations, the improvements of using n-n alignments versus 1-1 alignments is confirmed in both human evaluation and extrinsic evaluation.

\subsection{Intrinsic Evaluation}
\label{sec:intrinsic_evaluation}
\paragraph{Predictive Model Evaluation}
We compare the CS point predictions provided by the predictive model against the actual CS points in the CS sentences in ArzEn-ST dev set. We present accuracy, precision, recall, and F1 scores in Table \ref{table:PM_intrinsic_performance}. While these figures give us an intuition on the performance of the predictive models, it is to be noted that false positives are not necessarily incorrect. It is also to be noted that the high accuracy values are due to the high rate of true negative predictions.

As another evaluation, we check the POS distribution of the words predicted as CS by both the random and predictive models, against that of CS words in ArzEn-ST dev set. The predictive model shows a higher correlation (0.984) versus random approach (0.938). The POS distribution of the top frequent tags is shown in Appendix~\ref{sec:appendix-POS}. The predictions of the learnt model are dominated by nouns, followed by verbs and adjectives, where other POS tags have lower frequencies than in ArzEn-ST. The random approach gives better coverage for POS tags, however, introduces higher frequencies for low-frequent POS tags of CS words in ArzEn-ST.


\begin{table}[t]
\centering
\setlength{\tabcolsep}{3pt}
\begin{tabular}{ | l | c  c  c  c |}
\hline
\multicolumn{1}{|c|}{\textbf{Model}}&\textbf{Accuracy}&\textbf{Precision}&\textbf{Recall}&\textbf{F1}\\\hline
Random & 77.1 & 18.8 & 21.0 & 0.198\\
Predictive & \textbf{91.9} & \textbf{76.6} & \textbf{57.4} & \textbf{0.656}\\\hline
\end{tabular}
\caption{Evaluating the performance of the predictive model on the code-switched sentences in ArzEn-ST dev set.
}
\label{table:PM_intrinsic_performance}
\end{table}

\begin{table}[t]
\centering
\setlength{\tabcolsep}{3pt}
\begin{tabular}{ | l | c  c  c  r |}
\hline
&\multicolumn{1}{c}{\textbf{\%En}} &&&\multicolumn{1}{c|}{\textbf{\%En}}\\
\multicolumn{1}{|c|}{\textbf{Model}}&\multicolumn{1}{c}{\textbf{(words)}} &\textbf{CMI}&\textbf{av.}$|$\textbf{CS}$|$&\multicolumn{1}{c|}{\textbf{(sent.)}}\\\hline
{\dict} & 21.1 & 0.23 & 1.2 &0.0\\
{\randmapint} & 19.9 & 0.22 & 1.14 &0.0\\
{\predmapint} & 16.7 & 0.22 & 1.23 &6.3\\
{\randmapgdf} & 27.7 & 0.25 & 2.26 &6.8\\
{\predmapgdf} & 28.9 & 0.26 & 2.84 &18.3\\\hline
ArzEn-ST & 18.6 & 0.19 & 1.88 &3.7\\\hline
\end{tabular}
\caption{Evaluating augmented sentences in terms of CS metrics against ArzEn-ST train set.}
\label{table:syntheticData_CSMetrics}
\end{table}

\paragraph{CS Synthetic Data Analysis}
We look into how similar the synthetic data is to naturally occurring CS sentences. In Table \ref{table:syntheticData_CSMetrics}, we evaluate the synthetic data in terms of the percentage of English words, the Code-Mixing Index (CMI) \cite{DG14}, the average length of CS segments, as well as the percentage of monolingual English sentences generated. We observe that using 1-1 alignments, the generated CS sentences are close to natural occurring CS sentences in ArzEn-ST in terms of CS metrics. Using n-n alignments, the amount of CS in the synthetic data increases considerably.


\begin{table*}[th!]
\centering
\setlength{\tabcolsep}{2pt}
\begin{tabular}{ | l | r | c | c  c | l  l | l  l |}
\cline{3-9}
\multicolumn{2}{c|}{} &\multicolumn{1}{c|}{\textbf{LM}}&\multicolumn{2}{c|}{\textbf{ASR}}&\multicolumn{2}{c|}{\textbf{MT}} &\multicolumn{2}{c|}{\textbf{\textbf{ST}}} \\\hline 
\multicolumn{1}{|c|}{\textbf{Model}}& \multicolumn{1}{c|}{$|$\textbf{Train}$|$} &\textbf{PPL\textsubscript{\textit{All}}} &\textbf{WER\textsubscript{\textit{All}}} & \textbf{CER\textsubscript{\textit{All}}} & \textbf{chrF++\textsubscript{\textit{All}}} &\textbf{chrF++\textsubscript{\textit{CS}}} &\textbf{chrF++\textsubscript{\textit{All}}} &\textbf{chrF++\textsubscript{\textit{CS}}}\\\hline
Baseline & & 415.1 & 34.7 & 20.0 & 53.0 & 54.0 & 39.4 & 40.4 \\
+\dict & +240,678 & 313.3 & 33.2 & 19.1 & 52.6 & 53.5 & 40.1 & 41.0 \\
+\randmapint & +240,869 & 306.1 & 32.9 & 19.0 & 55.2$^{\ast}$ & 57.0$^{\ast}$ & 41.0$^{\ast}$ & 42.1$^{\ast}$\\
+\predmapint & +177,633 & 273.4 & 33.2 & 19.1 & 55.5$^{\dagger}$ & 57.4$^{\dagger}$ & 40.9$^{\dagger}$ & 42.2$^{\dagger}$ \\
+\randmapgdf & +207,026 & 273.8 & \textbf{32.9} & \textbf{18.9} & \textbf{56.0}$^{\ast}$ & \textbf{57.9}$^{\ast}$ & 41.4$^{\ast}$ & 42.7$^{\ast}$ \\
+\predmapgdf & +138,544 & 274.5 & 33.0 & \textbf{18.9} & \textbf{56.0}$^{\dagger}$ & 57.8$^{\dagger}$ & \textbf{41.5}$^{\dagger}$ & \textbf{42.8}$^{\dagger}$\\
+ExtraCS &  & 228.1 & 33.3 & 19.0 & 55.7 & 57.6 & 41.6 & 42.9 \\
\hline
\hline
\multicolumn{9}{|c|}{\textbf{Constrained Experiments}}\\\hline
+c[\dict] & +99,725 & 324.2 & 33.5 & 19.3 & 52.3 & 53.3 & 39.4 & 40.1 \\
+c[\randmapgdf] & +99,725 & 293.4 & 33.1 & 19.0 & 55.6$^{\star}$ & 57.3$^{\star}$ & \underline{41.2} & \underline{42.6} \\
+c[\predmapgdf] & +99,725 & \underline{270.4} & \underline{33.0} & \underline{18.9} & \underline{56.0}$^{\star}$ & \underline{57.9}$^{\star}$ & \underline{41.2} & \underline{42.6} \\\hline
\end{tabular}
\caption{We report the results of the extrinsic tasks on ArzEn-ST test set. For language modeling, we report PPL on all sentences. For ASR, we report WER and CER on all sentences. For MT and ST, we report chrF++ on all and CS sentences. We report the results of using all augmentations (non-constrained), followed by the constrained experiments.  The best performing approach in the non-constrained setting is bolded. The best performing approach in the constrained setting is underlined. We run statistical significance tests between {\randmap} and {\predmap} as well as 1-1 and n-n experiments, and mark models that are statistically significant (\textit{p-values}$<0.05$) with superscript symbols ($\ast,\dagger,\star$).}
\label{table:extrinsic_eval_results}
\end{table*}

\subsection{Extrinsic Evaluation}
We evaluate the improvements achieved through data augmentation on LM, ASR, MT, and ST tasks. Results are shown in Table \ref{table:extrinsic_eval_results}. We present perplexity (PPL) for LM and Word Error Rate (WER) and Character Error Rate (CER) for ASR. For MT and ST, we use BLEU \cite{PRT+02}, chrF, chrF++ \cite{Pop17}, and BERTScore (F1) \cite{ZKW+19}. 
BLEU, chrF and chrF++ are calculated using SacrebleuBLEU \cite{Pos18}. In Table \ref{table:extrinsic_eval_results}, we present the chrF++ scores. We present the results for all metrics in Appendix \ref{sec:appendix_mt_results}. 

\paragraph{Language Modeling}

PPL reductions are observed when using n-n over 1-1 alignments for random-based replacements. While {\randmapgdf} generates more data than {\predmapgdf}, both approaches achieve similar PPL, outperforming {\dict}. 
Overall, we achieve a 34\% reduction in PPL over baseline. 

\paragraph{ASR}
All models utilizing augmented data outperform the baseline. The best results are achieved using {\predmapgdf} and {\randmapgdf}, which perform equally well, achieving 5.2\% absolute WER reduction over baseline. 
We observe that these models 
slightly outperform 
those trained on extra real CS data.\footnote{It is to be noted that the data collected from social media platforms is noisy, however, it still brings improvements in LM and ASR tasks.}

\paragraph{Machine Translation Evaluation}

Results show that using n-n alignments outperforms 1-1 alignments on all settings. However, using a predictive model does not outperform random replacements. We observe that dictionary-based replacement negatively affects the MT systems. 
We also observe that our top two models perform equally well as the model utilizing real CS data, confirming the effectiveness of data augmentation, achieving $3$-$3.9$ chrF++ points over the baseline.

\paragraph{MT Qualitative Analysis}
When looking into the translations provided by the baseline model, we observe that many CS words get dropped in translation or get mistranslated. 
When checking the translations provided by the MT systems trained using augmentations, we observe that the majority of the CS words are retained through translation. We also observe that these MT systems are able to retain CS OOV words, where the words are not available in the baseline training data, nor introduced in the synthetic data. This shows that by adding CS synthetic sentences to the training set, 
the models learn to retain English words in translation. Examples are shown in Appendix~\ref{sec:appendix-translationExamples}.
	

\begin{figure}[t]
    \centering
    \includegraphics[width=\columnwidth]{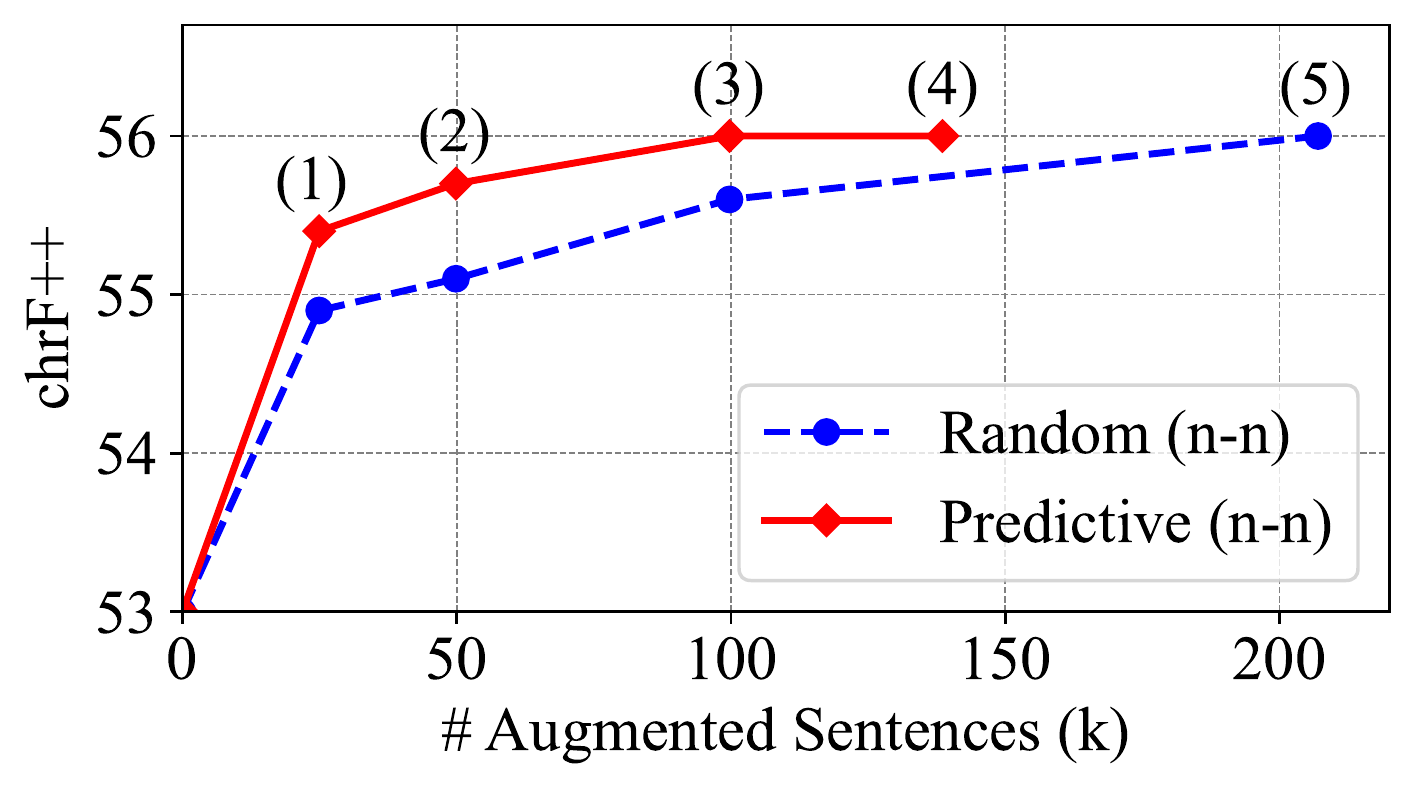}
    \caption{The chrF++ scores reported on ArzEn-ST test set when adding: (1) 25\% of the sentences in the constrained experiment (=$24.9k$), (2) 50\% of the sentences in the constrained experiment (=$49.8k$), (3) 100\% of the sentences in the constrained experiment (=$99.7k$), (4) all sentences generated by \predmapgdf~(=$138.5k$), and (5) all sentences generated by \randmapgdf~(=$207k$).}
    \label{fig:learning_curve}
\end{figure}

\paragraph{Speech Translation Evaluation}
Similar to previous results, both {\predmap} and {\randmap} outperform \dict. We observe improvements for using n-n alignments over using 1-1 alignments. However, no improvements are achieved by using predictive model over random predictions.

\begin{table*}[th!]
\centering
\setlength{\tabcolsep}{3pt}
\begin{tabular}{ | c | l | }
\hline
\multicolumn{2}{|c|}{\textbf{Understandability}}\\\hline
1  & No, this sentence doesn't make sense. \\
2  & Not sure, but I can guess the meaning of this sentence. \\
3 &  Certainly, I get the meaning of this sentence. \\\hline 
\multicolumn{2}{|c|}{\textbf{Naturalness}}\\\hline
1 & Unnatural, and I can't imagine people using this style of code-mixed Arabic-English. \\
2 & Weird, but who knows, it could be some style of code-mixed Arabic-English.\\
3 & Quite natural, but I think this style of code-mixed Arabic-English is rare.\\ 
4 & Natural, and I think this style of code-mixed Arabic-English is used in real life.\\
5 & Perfectly natural, and I think this style of code-mixed Arabic-English is very frequently used.\\\hline
\end{tabular}
\caption{The evaluation dimensions for human evaluation, following \cite{PC21}.}
\label{table:human_eval_rubrics}
\end{table*}

\paragraph{Constrained Experiments} In order to control existing variables, such as the number of generated sentences, and how similar they are to the test set, we conduct further experiments where we restrict the augmented sentences in each approach to the CS sentences that are generated across the three techniques: \dict, \randmapgdf, and \predmapgdf. We report results by training our models using these restricted augmentations ($99.7k$ sentences) in addition to the baseline training data in Table \ref{table:extrinsic_eval_results}. We find that, under this condition, for the MT task, the predictive model outperforms random, where the improvements are statistically significant on BLEU, chrF, and chrF++, as shown in Table \ref{table:mt_eval_all}. For the ASR task, while \predmapgdf~achieves lower PPL over \randmapgdf, both 
models perform equally. In Figure \ref{fig:learning_curve}, we show the learning curves for \randmapgdf~and \predmapgdf~MT scores when including 25\%, 50\%, and 100\% of the generated sentences in the constrained setting, in addition to the scores of the non-constrained setting. We see that \predmapgdf~ achieves overall the same performance as \randmapgdf~with half the amount of generated sentences.

\subsection{Human Evaluation}
We perform a human evaluation study to assess the quality of sentences generated by the five models: \randmapint, \predmapint, \randmapgdf, \predmapgdf, and \dict. 
Out of the sentences that get augmented in all five techniques, we randomly sample 150 sentences, and ask human annotators to judge the synthetic sentences generated by each model, giving a total of 750 sentences to be evaluated.\footnote{The sentences are sampled uniformly across the 6 corpora used in data augmentation to have equal representation of the different data sources (web/chat/conversational).} 
We also include 150 random CS sentences from ArzEn-ST to act as control sentences. These 900 sentences were judged by three bilingual Egyptian Arabic-English speakers. Following \cite{PC21}, the sentences are evaluated against understandability and naturalness, where the rubrics are outlined in Table \ref{table:human_eval_rubrics}.

For each synthetic/real sentence, we calculate the mean opinion score (MOS), which is the average of the three annotators' scores for that sentence. In Table \ref{table:MOS_scores}, we present the MOS distribution for each augmentation approach, presenting the percentage of sentences falling in each evaluation range. We observe that the annotators prefer the synthetic data generated using segment replacements (n-n alignments) over those using word replacements (1-1 alignments). The annotators also prefer the synthetic data generated using trained predictive models over those using random CS point prediction. The highest scores are achieved by \randmapgdf, where 44\% of the synthetic sentences are perceived as natural. 

\begin{table}[t]
\centering
\setlength{\tabcolsep}{1pt}
\begin{tabular}{ | l | r  r   r   r  r  r | }
\hline
\multicolumn{1}{|c|}{}&\multicolumn{1}{c}{}&\multicolumn{1}{c}{}&
\multicolumn{1}{c}{\sc \textbf{Rand}}&
\multicolumn{1}{c}{\sc \textbf{Pred}}&
\multicolumn{1}{c}{\sc \textbf{Rand}}&
\multicolumn{1}{c|}{\sc \textbf{Pred}}\\
\multicolumn{1}{|c|}{\textbf{MOS}}&\multicolumn{1}{c}{\textbf{ArzEn}}&\multicolumn{1}{c}{\sc \textbf{dict}}&
\multicolumn{1}{c}{\textbf{(1-1)}}&
\multicolumn{1}{c}{\textbf{(1-1)}}&
\multicolumn{1}{c}{\textbf{(n-n)}}&
\multicolumn{1}{c|}{\textbf{(n-n)}}\\\hline
\multicolumn{7}{|c|}{\textbf{Understandability}}\\\hline
1$\leq$*$\<2$&2.7&62.0&32.7&32.0&21.3&16.7\\
2$\leq$*$\<3$&97.3 & 38.0 & 67.3 & 68.0 & 78.7 & 83.3\\\hline
\multicolumn{7}{|c|}{\textbf{Naturalness}}\\\hline
1$\leq$*$\<2$&0.7&82.7&70.7&50.0&46.7&30.0\\
2$\leq$*$\<3$&6.0&8.7&12.7&18.0&26.0&25.3\\
3$\leq$*$\<4$&11.3&6.0&8.0&20.0&14.0&26.0\\
4$\leq$*$\leq5$&82.0&2.7&8.7&12.0&13.3&18.7\\\hline
\end{tabular}
\caption{The mean opinion score (MOS) distribution for synthetic sentences, showing the percentage of sentences falling in each evaluation range.
}
\label{table:MOS_scores}
\end{table}	

\section{Discussion}
\label{sec:discussion}
In this section, we revisit our RQs:

\paragraph{RQ1 - Can a model learn to predict CS points using limited amount of CS data?}
As shown in the intrinsic evaluation, the model learns to predict CS points to some extent, as shown in the improvements in accuracy, precision, and F1 scores over random predictions. 
This is also observed where the POS distribution of the CS predictions using a predictive model has higher correlation to the distribution found in natural CS sentences compared to random predictions.

\paragraph{RQ2 - Can this information be used to generate more natural synthetic CS data?}
Yes, this was confirmed though human evaluation, where annotators reported higher scores for understandability and naturalness using the predictive model over using random replacements.

\paragraph{RQ3 - Would higher quality of synthesized CS data necessarily reflect in performance improvements in downstream tasks?}
In the scope of our experiments, such an entailment does not necessarily hold. We believe two limitations are affecting the performance of the predictive model. First of all, 
the {\predmap} approach is based on the assumption that the data provided to the predictive model is representative enough of the CS phenomenon and includes all CS patterns. Due to the scarcity of CS corpora and the dynamic behaviour of CS  \cite{el2020and}, this point presents a challenge and could be restricting the potential power of this model, and it could be the case that {\randmap} is able to cover more CS patterns. This is supported by the POS distribution analysis in Section \ref{sec:intrinsic_evaluation}. 
Secondly, random has the power of generating more data as opposed to using a predictive model. 
When we control for size, we observe improvements in MT using the predictive model. In the future, we plan to work on improving the predictive approach to generate more CS sentences. 
For ASR, both approaches perform equally. It was also shown in \cite{hussein2023textual} that random lexical replacement outperforms the use of Equivalence Constraint linguistic theorem for ASR. Therefore, we believe further research is needed to draw strong conclusions about the relation between the quality of generated CS data and the improvements on different downstream tasks.

\section{Conclusion and Future Work}
\label{sec:conclusion}
In this paper, we investigate data augmentation for CS Egyptian Arabic-English. We utilize parallel corpora to perform lexical replacements, where CS points are either selected randomly or based on predictions of a neural-based model that is trained on a limited amount of CS data. We investigate word replacements using intersection alignments as well as segment replacements using symmetrized alignments. We compare both aligned-based replacements with dictionary-based replacements. We evaluate the effectiveness of data augmentation on LM, MT, ASR, and ST tasks, as well as assess the quality through human evaluation. Across all evaluations, we report that segment replacements outperform word replacements, and aligned-based replacements outperform dictionary-based replacements. The human evaluation study shows that utilizing predictive models produces augmented data of highest quality. For the downstream tasks, random and predictive techniques achieve similar results, both outperforming dictionary-based replacements. We observe that random has the advantage of generating more data. When controlling for the amount of generated data, the predictive technique outperforms random on the MT task. Our best models achieve 34\% improvement in perplexity, 5.2\% relative improvement on WER for ASR task, +4.0-5.1 BLEU points on MT task, and +2.1-2.2 BLEU points on ST task.



\section*{Acknowledgements}
We would like to thank Bashar Alhafni for the helpful discussions and the reviewers for their insightful comments. This project has benefited from financial support by DAAD (German Academic Exchange Service). 

\bibliography{references}
\bibliographystyle{acl_natbib}

\appendix
\section*{Limitations}
To the best of our knowledge, this paper presents the first comparison for the mentioned lexical replacement techniques, covering human evaluation as well as three downstream tasks; automatic speech recognition, machine translation, and speech translation. However, the study is focused on the Egyptian Arabic-English language pair, and we make no assumptions on the generalizability of results to other language pairs, nor other domains. Further investigations are needed to assess how the results would differ, especially in the case of languages with less syntactic divergence. We also note another limitation in the human evaluation, which is that code-switching is a user-dependent behaviour, that differs across different users, and thus the evaluation of the naturalness of a code-switched sentence is very subjective. We have taken this into account in our human evaluation study by having each sentence evaluated by three annotators and taking the average across the three ratings.
\section*{Ethics Statement}
We could not identify potential harm from using the provided models in this work. However, one concern is that code-switched ST is yet a challenging task, and the ST models trained in this work provide low performance, and thus should not be deployed as it can mislead the users.
\section{ArzEn-ST Corpus}
\label{sec:appendix-arzenExamples}
In Table \ref{table:arzen}, we provide an overview on ArzEn-ST corpus. In Table \ref{table:arzen_examples}, we show examples from the corpus.

\begin{table}[h]
\centering
\begin{tabular}{|p{5cm} | r |}
      \hline
      \multicolumn{2}{|c|}{\textbf{ArzEn-ST Speech Corpus}} \\\hline
      Duration&12h\\
      \#Speakers&40\\
      \# Sentences&6,216\\
      \% CS sentences&63.7\%\\
      \% Arabic sentences&33.2\%\\
      \% English sentences&3.1\%\\
      \hline
\end{tabular}
\caption{ArzEn-ST corpus overview.}
\label{table:arzen}
\end{table}

\begin{table}[h]
\centering
\setlength{\tabcolsep}{2pt}
\begin{tabular}{|l|r|}
\hline
\textbf{\#}&\multicolumn{1}{c|}{\textbf{Example}}\\\hline
&
project code
\<انا كتبت ال>$\leftarrow$\\
1&\multicolumn{1}{l|}{{\it AnA ktbt Al} project code}\\
&\multicolumn{1}{l|}{I wrote the project code}\\\hline
&
internship
\<عملت كذا>$\leftarrow$\\
2&\multicolumn{1}{l|}{{\it Emlt k*A} internship}\\
&\multicolumn{1}{l|}{I did several internships}\\\hline
&
\<الناس اللى معايا>
overload 
\<كنت ب>$\leftarrow$\\
3&\multicolumn{1}{l|}{{\it knt b} overload {\it AlnAs Ally mEAyA}}\\
&\multicolumn{1}{l|}{I was overloading my teammates}\\\hline
&\<معينة>
traffic within period 
\<ال>
detect   
\<ن>$\leftarrow$\\
4&\multicolumn{1}{l|}{{\it n} detect {\it Al} traffic within period {\it mEynp}}\\
&\multicolumn{1}{l|}{to detect the traffic within a certain period}\\\hline
\end{tabular}
\caption{ArzEn-ST corpus examples, showing source text, its transliteration \cite{Habash:2007:arabic-transliteration},
and translation. The arrows beside the sentences show the sentence starting direction, as Arabic is read right to left.}
\label{table:arzen_examples}
\end{table}

\section{POS Intrinsic Evaluation}
\label{sec:appendix-POS}
\begin{table}[]
\centering
\begin{tabular}{ |l | r  r  r |}
\hline
\multicolumn{1}{|c|}{\textbf{POS}} & \multicolumn{1}{c}{\textbf{ArzEn}} & \multicolumn{1}{c}{\textbf{Random}} & \multicolumn{1}{c|}{\textbf{Predictive}} \\
\hline
NN & 48.4 & 33.2 & 67.0 \\
VB & 14.5 & 22.9 & 13.6 \\
JJ & 13.1 & 9.3 & 13.6 \\
RB & 7.6 & 6.5 & 1.7 \\
IN & 5.0 & 8.9 & 0.9 \\
PRP & 3.8 & 4.7 & 0.6 \\
DT & 2.2 & 3.7 & 0.2 \\
CC & 0.9 & 3.8 & 0.1 \\\hline
Total & 94.7 & 89.3 & 97.6\\\hline
\end{tabular}
\caption{The POS distribution (\%) of the words predicted as CS words by both the random and predictive models, against that of CS words in ArzEn-ST dev set.}
\label{table:pos_tags}
\end{table}
As an intrinsic evaluation of the CS predictive model, we check the POS distribution of the words predicted as CS words by both the random and predictive approaches, against that of CS words in ArzEn-ST dev set. We report that the natural POS distribution is in-line with the distributions reported for CS Egyptian Arabic-English \cite{HEA18,BHA+20}, where the dominating POS tags are nouns, verbs, and adjectives, followed by adverbs, pronouns, and prepositions. We report that the predictive model gives a higher correlation (0.984) versus random approach (0.938). We present the POS distribution of the top frequent tags in Table~\ref{table:pos_tags}. 
We observe that the predictive model provides a percentage of nouns that is significantly higher than that occurring in ArzEn-ST. It also provides less coverage to the tags occurring less frequently in ArzEn-ST. 
We believe this can be due to the predictive model being trained on limited data. The random approach on the other hand, provides higher counts for less frequent POS tags, as seen in the total, where 11\% of the words identified by the random prediction to be code-switched belong to POS tags that are infrequent in natural CS data.

\section{Data Preprocessing}
\label{sec:appendix_data_preprocessing}
Data preprocessing involved removing corpus-specific annotations, removing URLs and emoticons through \textit{tweet-preprocessor},\footnote{\url{https://pypi.org/project/tweet-preprocessor/}} tokenizing numbers, lowercasing, running Moses’ \cite{KHB+07} tokenizer as well as MADAMIRA \cite{PAD+14} simple tokenization (D0), and performing Alef/Ya normalization. For LDC2017T07 \cite{LDC2017T07}, LDC2019T01 \cite{LDC2019T01}, and LDC2021T15 \cite{LDC2021T15}, 
some words have literal and intended translations. We opt for one translation having all literal translations and another having all intended translations. For LDC2017T07, we utilize the work by \newcite{SUH20}, where the authors used a sequence-to-sequence deep learning model to transliterate SMS/chat text in LDC2017T07 from Arabizi (where Arabic words are written in Roman script) to Arabic orthography.

\begin{table*}[t]
\centering
\setlength{\tabcolsep}{2pt}
\begin{tabular}{ | l | l  l  l  c  c | l  l  l  c  c |}
\cline{2-6}\cline{7-11}
\multicolumn{1}{c}{}&\multicolumn{5}{|c|}{\textbf{All Sentences}}&\multicolumn{5}{c|}{\textbf{CS Sentences}}\\\hline
\multicolumn{1}{|c|}{\textbf{Model}}&\textbf{BLEU}&\textbf{chrF}&\textbf{chrF++}&\textbf{F\textsubscript{\textit{BERT}}}&\textbf{Avg\textsubscript{\textit{MT}}}&\textbf{BLEU}&\textbf{chrF}&\textbf{chrF++}&
\textbf{F\textsubscript{\textit{BERT}}} &\textbf{Avg\textsubscript{\textit{MT}}} \\\hline\hline
\multicolumn{11}{|c|}{\textbf{Non-constrained Experiments}}\\\hline
\multicolumn{11}{|c|}{\textbf{MT}}\\\hline
Baseline&31.0&54.2&53.0&0.519&47.5&31.4&55.3&54.0&0.501&47.7\\
+\dict&30.9&53.8&52.6&0.516&47.2&31.5&54.7&53.5&0.498&47.4\\
+\randmapint&34.4$^{\ddagger}$&56.6$^{\ast}$&55.2$^{\ast}$&0.545&50.2&35.9$^{\ddagger}$&58.5$^{\ast,\ddagger}$&57.0$^{\ast}$&0.543&51.4\\
+\predmapint&33.7$^{\ddagger,\dagger}$&56.9$^{\dagger}$&55.5$^{\dagger}$&0.548&50.2&35.2$^{\ddagger,\dagger}$&58.9$^{\dagger,\ddagger}$&57.4$^{\dagger}$&0.549&51.6\\
+\randmapgdf&34.7&57.2$^{\ast}$&\textbf{56.0}$^{\ast}$&\textbf{0.552}&\textbf{50.8}&36.2&\textbf{59.2}$^{\ast}$&\textbf{57.9}$^{\ast}$&\textbf{0.552}&\textbf{52.1}\\
+\predmapgdf&\textbf{35.0}$^{\dagger}$&\textbf{57.3}$^{\dagger}$&\textbf{56.0}$^{\dagger}$&0.550&\textbf{50.8}&\textbf{36.5}$^{\dagger}$&\textbf{59.2}$^{\dagger}$&57.8$^{\dagger}$&\textbf{0.552}&\textbf{52.1}\\
+ExtraCS&34.8&57.2&55.7&0.547&50.6&36.2&59.1&57.6&0.546&51.9\\\hline
\multicolumn{11}{|c|}{\textbf{ST}}\\\hline
Baseline&15.3&41.2&39.4&0.335&32.4&15.8&42.4&40.4&0.317324&32.6\\
+\dict&16.3&41.9&40.1&0.344&33.2&16.8&42.8&41.0&0.324&33.2\\
+\randmapint&16.5$^{\ddagger,\ast}$&42.8$^{\ast}$&41.0$^{\ast}$&0.347&33.8&17.0$^{\ast}$&44.1$^{\ast}$&42.1$^{\ast}$&0.329&34.0\\
+\predmapint&16.1$^{\ddagger,\dagger}$&42.8$^{\dagger}$&40.9$^{\dagger}$&0.348&33.6&16.9$^{\dagger}$&44.2$^{\dagger}$&42.2$^{\dagger}$&0.331&34.1\\
+\randmapgdf&\textbf{17.0}$^{\ast}$&43.3$^{\ast}$&41.4$^{\ast}$&0.349&\textbf{34.2}&\textbf{17.7}$^{\ast}$&44.7$^{\ast}$&42.7$^{\ast}$&0.332&\textbf{34.6}\\
+\predmapgdf&16.9$^{\dagger}$&\textbf{43.4}$^{\dagger}$&\textbf{41.5}$^{\dagger}$&\textbf{0.352}&\textbf{34.2}&17.4$^{\dagger}$&\textbf{44.8}$^{\dagger}$&\textbf{42.8}$^{\dagger}$&\textbf{0.335}&\textbf{34.6}\\
+ExtraCS&17.4&43.4&41.6&0.353&34.4&18.0&44.7&42.9&0.336&34.8\\\hline
\hline
\multicolumn{11}{|c|}{\textbf{Constrained Experiments}}\\\hline
\multicolumn{11}{|c|}{\textbf{MT}}\\\hline
+c[\dict]&30.3&53.6&52.3&0.517&47.0&31.0&54.6&53.3&0.499&47.2\\		
+c[\randmapgdf]&33.8$^{\star}$&56.9$^{\star}$&55.6$^{\star}$&\underline{0.553}&50.4&35.1$^{\star}$&58.7$^{\star}$&57.3$^{\star}$&\underline{0.555}&51.7\\
+c[\predmapgdf]&\underline{35.0}$^{\star}$&\underline{57.4}$^{\star}$&\underline{56.0}$^{\star}$&0.551&\underline{50.9}&\underline{36.8}$^{\star}$&\underline{59.5}$^{\star}$&\underline{57.9}$^{\star}$&0.554&\underline{52.4}\\\hline
\multicolumn{11}{|c|}{\textbf{ST}}\\\hline
+c[\dict]&15.2&41.2&39.4&0.341&32.5&15.5&42.1&40.1&0.319&32.4\\	
+c[\randmapgdf]&16.4&\underline{43.1}&\underline{41.2}&0.350&33.9&17.0&\underline{44.7}&\underline{42.6}&0.335&\underline{34.5}\\
+c[\predmapgdf]&\underline{16.6}&\underline{43.1}&\underline{41.2}&\underline{0.353}&\underline{34.0}&\underline{17.2}&44.6&\underline{42.6}&\underline{0.337}&\underline{34.5}\\\hline
\end{tabular}
\caption{MT and ST evaluation on ArzEn-ST test set for the non-constrained (using all augmentations) and constrained experiments. We report BLEU, chrF, chrF++, F1 BERTScore (F$_{BERT}$), and their average (Avg$_{MT}$), on all sentences as well as code-switched sentences only. The best performing  data augmentation approach in the non-constrained setting is bolded. The best performing approach in the constrained setting is underlined. We run statistical significance tests between pairs of models to compare the effect of using {\randmap} vs. {\predmap} and 1-1 vs. n-n alignments, and mark models that are statistically significant (\textit{p-values}$<0.05$) with superscript symbols ($\ast,\dagger,\ddagger$,$\star$).
}
\label{table:mt_eval_all}
\end{table*}

\section{MT Hyperparameters}
\label{sec:appendix-MThyperparameters}
The following is the train command:\newline
python3 fairseq\_cli/train.py \${DATA\_DIR} --source-lang src --target-lang tgt --arch transformer --share-all-embeddings --encoder-layers 5 --decoder-layers 5 --encoder-embed-dim 512 --decoder-embed-dim 512 --encoder-ffn-embed-dim 2048 --decoder-ffn-embed-dim 2048 --encoder-attention-heads 2 --decoder-attention-heads 2 --encoder-normalize-before --decoder-normalize-before  --dropout 0.4 --attention-dropout 0.2 --relu-dropout 0.2  --weight-decay 0.0001 --label-smoothing 0.2 --criterion label\_smoothed\_cross\_entropy --optimizer adam --adam-betas '(0.9, 0.98)' --clip-norm 0 --lr-scheduler inverse\_sqrt --warmup-updates 4000 --warmup-init-lr 1e-7  --lr 1e-3 --stop-min-lr 1e-9 --max-tokens 4000 --update-freq 4 --max-epoch 100 --save-interval 10 --ddp-backend=no\_c10d

\section{MT Results}
\label{sec:appendix_mt_results}
In Table \ref{table:mt_eval_all}, we present the MT and ST results of the non-constrained and constrained experiments. We report the scores on BLEU, chrF, chrF++, and BERTScore(F1). Given that each metric has its strengths and weaknesses, we also report the average of the four metrics ($AvgMT$).



\section{Translation Examples}
\label{sec:appendix-translationExamples}
In Table \ref{table:translation_examples}, we show examples of source-target pairs with their translations obtained from different MT models. We observe that the models trained using augmented sentences are better than the baseline MT model at retaining CS words in the source sentence in the translations.
\definecolor{dark_green}{rgb}{0.0, 0.5, 0.0}
\definecolor{dark_orange}{rgb}{0.8, 0.33, 0.0}

\begin{table*}[t]
\centering
\setlength{\tabcolsep}{2pt}
\begin{tabular}{|l|p{12cm}|}
\hline
\multicolumn{1}{|c|}{\textbf{Model}}&\multicolumn{1}{c|}{\textbf{Example}}\\\hline
Src&
\multicolumn{1}{r|}{
\<فهو >
poker face
\<يببقوا>
adjudicate
\<ما هو المفروض ال . . ال . . الناس اللي بت>
}
\\
&
\multicolumn{1}{r|}{
\<ماينفعش يفهمني اي حاجة بس بعديها بيبقي يعني بعرف غلطي , بس>
}\\
Tgt-Ref & those one who \underline{adjudicate} should have a \underline{poker face}, so i can't get any signal from them, but afterwards i know my mistake, that's all\\
Baseline&it's supposed to be the. the. people who are \underline{a rijudi} could be \underline{powder face} so it can't explain anything but after that i mean i know my mistake, that's it\\
\dict&the.. the. the.. the people who \underline{are hurt} should be \underline{thinking about face}, so it can't explain anything to me, but after that, i mean, i know my mistake, that's it\\
\randmapint&the.. the.. the.. the.. the people who \underline{adjudicate} become a \underline{poker of face}, so he can't explain anything to me, but after that i know my mistake, that's it\\
\predmapint&the.. the. the.. the people that \underline{adjudicate} become the \underline{poker face}, so he can't understand anything but after that i mean i know my mistake, that's it\\
\randmapgdf&the.. the.. the.. the people who are \underline{adjudicate}, they become \underline{poker face}, so it can't explain anything to me after that, i mean, i know my mistake, that's it\\
\predmapgdf&the.. the. the.. the people who are \underline{adjudicate} should be \underline{poker face}, so he can't explain anything to me but after that, i mean, i know my mistake, that's it\\\hline
Src&
\multicolumn{1}{r|}{
multi-robot system task allocation
\<انا بعمل مشروع اسمه>}\\
Tgt-Ref&i'm working on a project called \underline{multi-robot system task allocation}.\\
Baseline&i make a project called \underline{multi-robot system and allocation}\\
\dict&i'm making a project called \underline{al-gamalt system for the task of allocation}\\
\randmapint&i make a project called \underline{multi-robot system and allocation}\\
\predmapint&i am making a project called \underline{multi-robot system and allocation task}\\
\randmapgdf&i am making a project called \underline{multi-robot system allocation}\\
\predmapgdf&i am doing a project called \underline{multi-robot system task allocation}\\\hline
\end{tabular}
\caption{Examples of translation outputs obtained from the MT models. The words in the translations that correspond to the CS words in the input source sentence are underlined. 
}
\label{table:translation_examples}
\end{table*}

\end{document}